\documentclass[12pt,a4paper]{article}

\usepackage[T1]{fontenc}
\usepackage[utf8]{inputenc}
\usepackage{lmodern}
\usepackage[margin=2.5cm]{geometry}
\usepackage{setspace}
\usepackage{booktabs}
\usepackage{array}
\usepackage{longtable}
\usepackage{tabularx}
\usepackage{multirow}
\usepackage{graphicx}
\usepackage{float}
\usepackage{caption}
\usepackage{subcaption}
\usepackage{amsmath}
\usepackage{amssymb}
\usepackage{microtype}
\usepackage{parskip}
\usepackage[hidelinks]{hyperref}
\usepackage{url}
\usepackage[numbers,square]{natbib}
\usepackage{authblk}
\usepackage{abstract}
\usepackage{titlesec}
\usepackage{xcolor}
\usepackage{colortbl}
\usepackage{rotating}
\usepackage{pdflscape}

\titleformat{\section}{\large\bfseries}{\thesection.}{0.5em}{}[\titlerule]
\titleformat{\subsection}{\normalsize\bfseries}{\thesubsection}{0.5em}{}

\definecolor{headerrow}{RGB}{50,50,50}
\definecolor{lightrow}{RGB}{245,247,250}
\definecolor{highlightrow}{RGB}{232,244,253}

\title{\textbf{Early Detection of Alzheimer's Disease Using Explainable
Machine Learning on Clinical Biomarkers: A Multi-Class Classification
Study Using the Alzheimer's Disease Neuroimaging Initiative (ADNI) Dataset}}

\author[1]{Afshan Hashmi}

\affil[1]{TRDC, Tuwaiq Academy,
Riyadh, Saudi Arabia\\
\small\texttt{afshanhashmi786@gmail.com};\quad
\texttt{a.hashmi@tuwaiq.edu.sa}}

\date{\today}

\begin{document}
\maketitle

\begin{abstract}
\noindent
\textbf{Background:} Alzheimer's disease (AD) affects over 55~million people
worldwide. Accurate, interpretable detection of normal cognition (NC), mild
cognitive impairment (MCI), and AD from routine clinical assessments remains
a critical unmet need.

\medskip
\noindent
\textbf{Methods:} An XGBoost classifier was developed for three-class detection
using eight clinical features from the Alzheimer's Disease Neuroimaging
Initiative (ADNI): MMSE, CDR Global, CDR Sum of Boxes (CDR-SB), MoCA, FAQ,
age, sex, and education. Hyperparameters were optimised using Optuna (50
trials); class imbalance was addressed with SMOTE. Performance was evaluated
by macro AUC-ROC with 1,000-iteration bootstrap 95\% confidence intervals,
macro F1, balanced accuracy, and Cohen's kappa. SHAP values provided
feature-level explainability.

\medskip
\noindent
\textbf{Results:} The dataset comprised 1,641 baseline subjects (608~NC,
767~MCI, 266~AD). On five-fold cross-validation, mean macro AUC was
0.983~(SD~0.007), accuracy 0.944~(SD~0.006), and macro F1 0.929~(SD~0.008).
On the held-out test set ($n=247$), macro AUC was 0.982 (95\% CI:
0.965--0.995), accuracy 0.943, balanced accuracy 0.932, macro F1~0.927, and
Cohen's $\kappa$~0.909. SHAP analysis identified CDR Global as the dominant
predictor for NC and MCI, while CDR-SB and MMSE together drove AD
classification.

\medskip
\noindent
\textbf{Conclusion:} An explainable machine learning model trained on routine
clinical assessments achieves near-perfect three-class Alzheimer's detection.
SHAP analysis reveals clinically plausible, class-specific feature importance
patterns supporting clinical validity. Future work will extend this framework
with speech biomarkers for multimodal detection.

\medskip
\noindent
\textbf{Keywords:} Alzheimer's disease; mild cognitive impairment; machine
learning; XGBoost; explainability; SHAP; ADNI; early detection; cognitive
assessment; gradient boosting
\end{abstract}

\section{Introduction}

Alzheimer's disease (AD) is the most prevalent neurodegenerative disorder
worldwide, accounting for 60--70\% of all dementia cases and affecting an
estimated 55~million people globally, a figure projected to exceed 139~million
by 2050~\cite{who2023,adi2023}. Despite its devastating clinical and
socioeconomic burden, no disease-modifying pharmacological intervention has
been approved, making early and accurate diagnosis a central clinical
priority~\cite{jack2018}.

Mild cognitive impairment (MCI) represents the transitional zone between
normal cognitive aging and manifest dementia, characterised by objective
memory decline that does not interfere substantially with daily
functioning~\cite{petersen2009}. Approximately 15\% of individuals diagnosed
with MCI progress to AD annually, with cumulative conversion rates reaching
30--40\% over three years~\cite{mitchell2009}. The accurate identification of
MCI patients at high conversion risk is therefore one of the most consequential
challenges in dementia research.

Standardised clinical instruments, including the Mini-Mental State Examination
(MMSE)~\cite{folstein1975}, Clinical Dementia Rating (CDR)~\cite{morris1993},
Montreal Cognitive Assessment (MoCA)~\cite{nasreddine2005}, and Functional
Activities Questionnaire (FAQ)~\cite{pfeffer1982}, form the cornerstone of
routine cognitive assessment. These instruments are widely deployed in memory
clinics globally, generate quantitative scores, and have demonstrated validity
in characterising cognitive status. However, their potential for automated,
multi-class simultaneous discrimination of NC, MCI, and AD within an
interpretable machine learning framework remains incompletely explored.

Gradient boosted decision tree models, particularly XGBoost~\cite{chen2016},
have demonstrated exceptional performance on heterogeneous tabular clinical
data, outperforming deep learning approaches in several medical prediction
benchmarks. Critically, unlike neural network models, gradient boosted trees
are compatible with SHAP (SHapley Additive exPlanations)~\cite{lundberg2017},
a theoretically grounded framework for computing feature-level contributions
to individual predictions. This combination of strong performance and
interpretability is essential for clinical adoption, where algorithmic
transparency is a regulatory and ethical requirement.

The Alzheimer's Disease Neuroimaging Initiative (ADNI)~\cite{weiner2017}
provides one of the largest, most comprehensively characterised multicentre
longitudinal datasets in dementia research. Several prior studies have used
ADNI for binary AD-versus-control classification~\cite{battineni2019,kavitha2022}
or MCI-to-AD conversion prediction~\cite{moradi2015,ritter2015}. However,
few have addressed three-class simultaneous discrimination of NC, MCI, and AD
using exclusively clinical assessment features, with rigorous external
validation and systematic per-class explainability analysis.

This study addresses this gap by developing and externally validating an
explainable XGBoost classifier for three-class Alzheimer's detection using
eight features from routine ADNI clinical assessments. Optuna-based Bayesian
hyperparameter optimisation, SMOTE for class imbalance correction, and SHAP
TreeExplainer for per-class feature importance analysis were applied. All
performance metrics are reported with bootstrap confidence intervals and
five-fold cross-validation in accordance with the updated TRIPOD+AI reporting
guidelines~\cite{collins2024}.

\section{Materials and Methods}

\subsection{Dataset and Ethical Considerations}

Data were obtained from the Alzheimer's Disease Neuroimaging Initiative (ADNI;
\url{adni.loni.usc.edu}). ADNI was launched in 2003 as a public-private
partnership, led by Principal Investigator Michael W.\ Weiner MD, with the
primary goal of testing whether serial magnetic resonance imaging, positron
emission tomography, other biological markers, and clinical and
neuropsychological assessment can be combined to measure the progression of
MCI and early AD. ADNI was approved by the institutional review boards of all
participating sites; all participants provided written informed consent. This
study used exclusively de-identified, publicly available data and was therefore
exempt from additional local ethical review.

\subsection{Subject Selection and Labelling}

Baseline visits (VISCODE~=~`bl') were used exclusively to simulate a
first-encounter clinical scenario and prevent temporal data leakage from
longitudinal follow-up. Subjects were classified into three diagnostic groups
based on the ADNI Diagnostic Summary: normal cognition (NC; DIAGNOSIS~=~1),
mild cognitive impairment (MCI; DIAGNOSIS~=~2), and Alzheimer's disease (AD;
DIAGNOSIS~=~3). This yielded 1,641 subjects: 608~NC (37.1\%), 767~MCI
(46.7\%), and 266~AD (16.2\%).

\subsection{Feature Extraction}

Eight features were extracted from five ADNI assessment tables at baseline:
(1)~MMSE total score~\cite{folstein1975};
(2)~CDR Global rating~\cite{morris1993};
(3)~CDR Sum of Boxes (CDR-SB)~\cite{obryant2008};
(4)~MoCA total score~\cite{nasreddine2005};
(5)~FAQ total score~\cite{pfeffer1982};
(6)~age derived from year of birth;
(7)~sex (binary encoded: male~=~1, female~=~0); and
(8)~years of formal education.
MoCA was missing for 59\% of subjects, FAQ for 1.1\%, and age for 0.1\%.
All missing values were imputed using median imputation fitted on the training
set only to prevent data leakage.

\subsection{Data Splitting and Class Imbalance}

Subjects were divided into training (70\%), validation (15\%), and test (15\%)
sets using stratified random sampling to preserve class proportions (random
seed~42). The held-out test set ($n=247$) was set aside prior to any model
development and used only for final evaluation. SMOTE~\cite{chawla2002} was
applied exclusively to the training set, generating synthetic minority-class
examples by interpolating between existing samples ($k=5$ neighbours), yielding
a balanced training set of 536~samples per class.

\subsection{Model Development and Hyperparameter Optimisation}

An XGBoost classifier~\cite{chen2016} was trained with the \texttt{multi:softprob}
objective for probabilistic three-class output. Hyperparameter optimisation was
performed using the Optuna framework~\cite{akiba2019} (50~trials,
Tree-structured Parzen Estimator sampler), minimising log loss on the
validation set. The search space included: number of estimators (200--800),
maximum tree depth (3--8), learning rate (0.01--0.2, log scale), subsample
ratio (0.6--1.0), column subsample ratio (0.6--1.0), and L1 regularisation
coefficient (0.0001--10.0, log scale). The optimal configuration was:
$n\_\text{estimators}=544$, $\text{max\_depth}=3$, $\text{lr}=0.199$,
$\text{subsample}=0.941$, $\text{colsample}=0.637$, $\alpha=0.134$.
Early stopping with patience of 30~rounds was applied during final training.

\subsection{Evaluation Metrics}

Model performance was evaluated using: (1)~macro AUC-ROC with 1,000-iteration
stratified bootstrap 95\% confidence intervals~\cite{delong1988};
(2)~overall accuracy; (3)~balanced accuracy; (4)~macro F1 score; and
(5)~Cohen's kappa~\cite{cohen1960}. Per-class sensitivity and specificity
were computed using one-versus-rest binarisation. Five-fold stratified
cross-validation was performed on the combined training and validation set
($n=1{,}394$).

\subsection{Explainability Analysis}

SHAP TreeExplainer~\cite{lundberg2017} was applied to all test set
predictions. Mean absolute SHAP values were calculated per feature per
diagnostic class, yielding a class-specific feature importance ranking. This
analysis captures the differential contribution of each clinical feature to
the model's discrimination of each diagnostic category.

\subsection{Reporting Standards}

This manuscript follows the TRIPOD+AI checklist~\cite{collins2024} for
transparent reporting. All analyses were performed in Python~3.12 using
XGBoost~v2.x, SHAP~v0.44, scikit-learn~v1.3, imbalanced-learn~v0.11, and
Optuna~v3.x. The complete source code and analysis pipeline are available at
\url{https://github.com/[to-be-added-upon-acceptance]}.

\section{Results}

\subsection{Dataset Characteristics}

\begin{figure}[H]
\centering
\includegraphics[width=\linewidth]{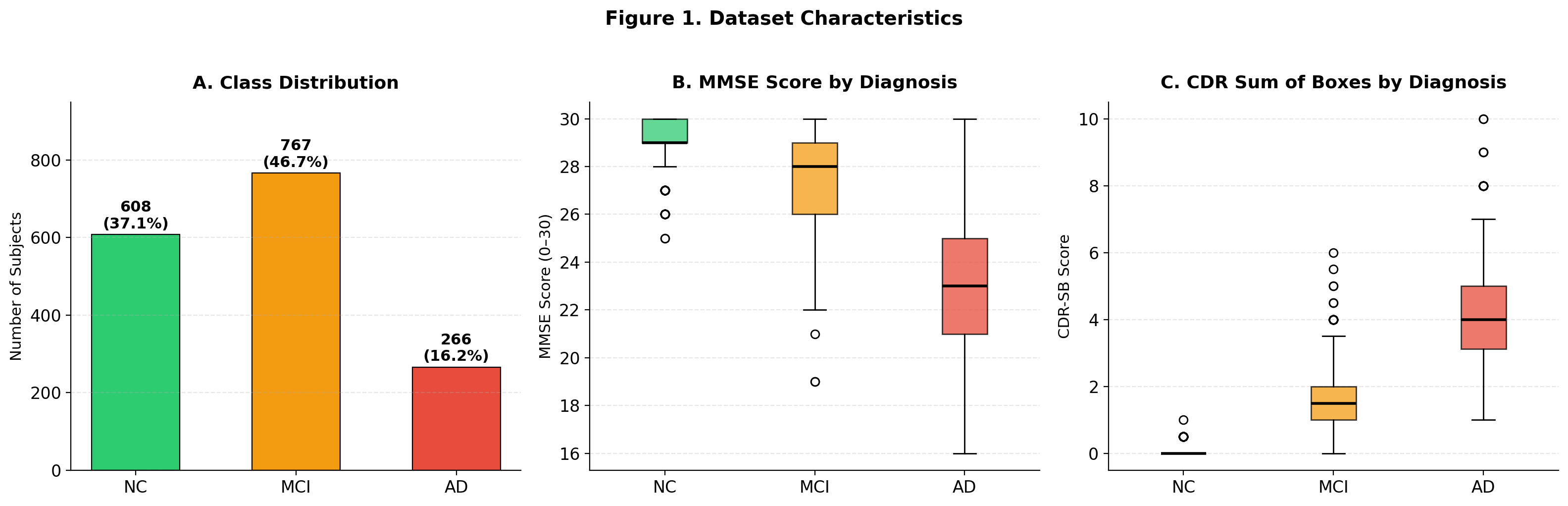}
\caption{Dataset characteristics. (A) Class distribution (NC~=~608, MCI~=~767, AD~=~266).
(B) MMSE score distribution by diagnosis group. (C) CDR Sum of Boxes distribution by
diagnosis group. Box plots show median, interquartile range, and 1.5$\times$~IQR whiskers.}
\label{fig:fig1}
\end{figure}

The final cohort comprised 1,641 subjects at baseline. Table~\ref{tab:table1}
presents clinical and demographic characteristics by diagnostic group.
Clear and statistically consistent differences were observed across all
features. NC subjects had the highest MMSE scores (29.1~$\pm$~1.0) and the
lowest CDR-SB (0.0~$\pm$~0.1) and FAQ scores (0.1~$\pm$~0.5). MCI subjects
showed intermediate profiles: MMSE 27.5~$\pm$~1.9, CDR-SB 1.5~$\pm$~0.9,
and FAQ 3.2~$\pm$~4.2. AD subjects demonstrated markedly impaired performance:
MMSE 23.2~$\pm$~2.3, CDR-SB 4.3~$\pm$~1.7, and FAQ 13.1~$\pm$~6.9. AD
subjects were older on average (73.2~$\pm$~10.1~years) compared to NC
(66.2~$\pm$~11.2~years). Figure~\ref{fig:fig1} illustrates class distribution
and score distributions by diagnosis group.

\begin{table}[H]
\centering
\caption{Baseline clinical and demographic characteristics by diagnostic group (mean $\pm$ SD).}
\label{tab:table1}
\rowcolors{2}{lightrow}{white}
\begin{tabular}{lrrr}
\toprule
\rowcolor{headerrow}
\textcolor{white}{\textbf{Variable}} &
\textcolor{white}{\textbf{NC ($n$=608)}} &
\textcolor{white}{\textbf{MCI ($n$=767)}} &
\textcolor{white}{\textbf{AD ($n$=266)}} \\
\midrule
Age (years)         & 66.2 $\pm$ 11.2 & 70.3 $\pm$ 10.4 & 73.2 $\pm$ 10.1 \\
Education (years)   & 16.5 $\pm$ 2.5  & 15.8 $\pm$ 2.8  & 15.0 $\pm$ 3.0  \\
MMSE score          & 29.1 $\pm$ 1.0  & 27.5 $\pm$ 1.9  & 23.2 $\pm$ 2.3  \\
CDR Global          & 0.0 $\pm$ 0.0   & 0.5 $\pm$ 0.1   & 1.0 $\pm$ 0.3   \\
CDR-SB              & 0.0 $\pm$ 0.1   & 1.5 $\pm$ 0.9   & 4.3 $\pm$ 1.7   \\
MoCA score          & 26.0 $\pm$ 2.7  & 22.7 $\pm$ 3.4  & 16.7 $\pm$ 4.3  \\
FAQ total           & 0.1 $\pm$ 0.5   & 3.2 $\pm$ 4.2   & 13.1 $\pm$ 6.9  \\
\bottomrule
\end{tabular}
\smallskip
\begin{minipage}{\linewidth}
\small\textit{NC = Normal Cognition; MCI = Mild Cognitive Impairment; AD = Alzheimer's Disease;
MMSE = Mini-Mental State Examination; CDR-SB = Clinical Dementia Rating Sum of Boxes;
MoCA = Montreal Cognitive Assessment; FAQ = Functional Activities Questionnaire.}
\end{minipage}
\end{table}

\subsection{Model Performance}

Table~\ref{tab:table2} presents model performance and compares it against
published studies. On five-fold cross-validation, the model achieved a mean
macro AUC of 0.983 (SD~0.007), ranging from 0.975 to 0.992 across folds.
Mean accuracy was 0.944 (SD~0.006) and macro F1 was 0.929 (SD~0.008).

\begin{figure}[H]
\centering
\includegraphics[width=\linewidth]{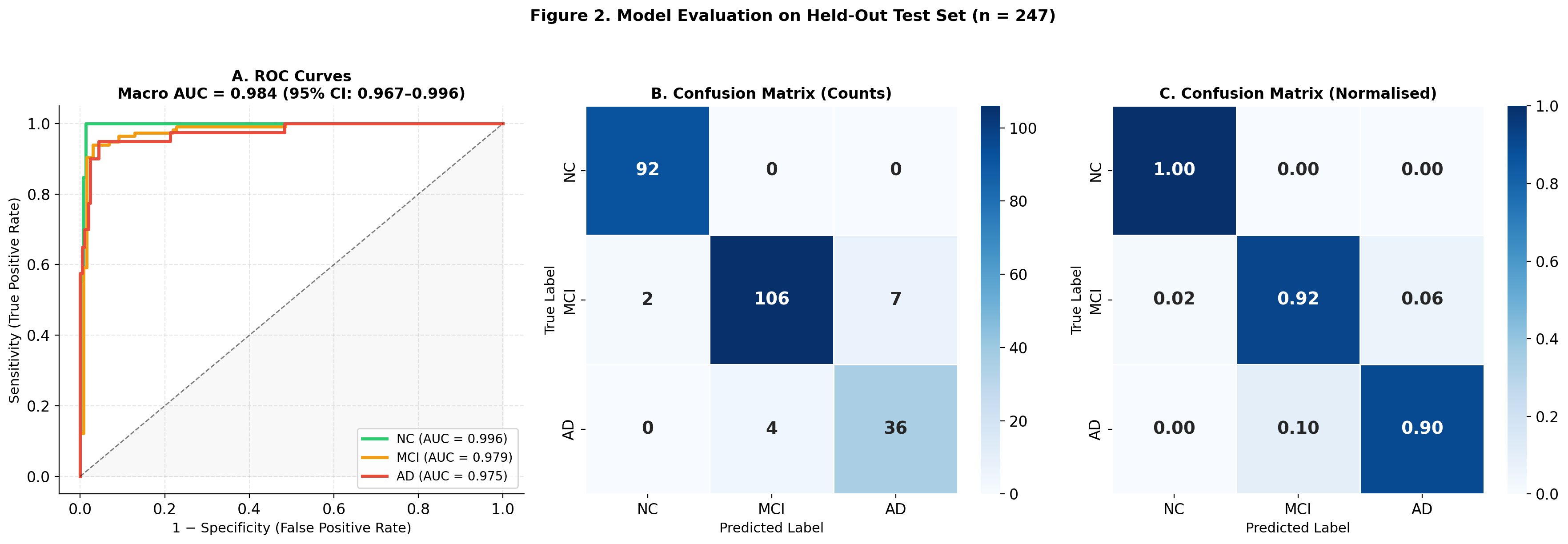}
\caption{Model evaluation on held-out test set ($n=247$). (A) Multi-class ROC curves with
per-class AUC. Macro AUC~=~0.982 (95\% CI: 0.965--0.995). (B) Confusion matrix (absolute
counts). (C) Normalised confusion matrix. NC~=~Normal Cognition; MCI~=~Mild Cognitive
Impairment; AD~=~Alzheimer's Disease.}
\label{fig:fig2}
\end{figure}

\begin{figure}[H]
\centering
\includegraphics[width=\linewidth]{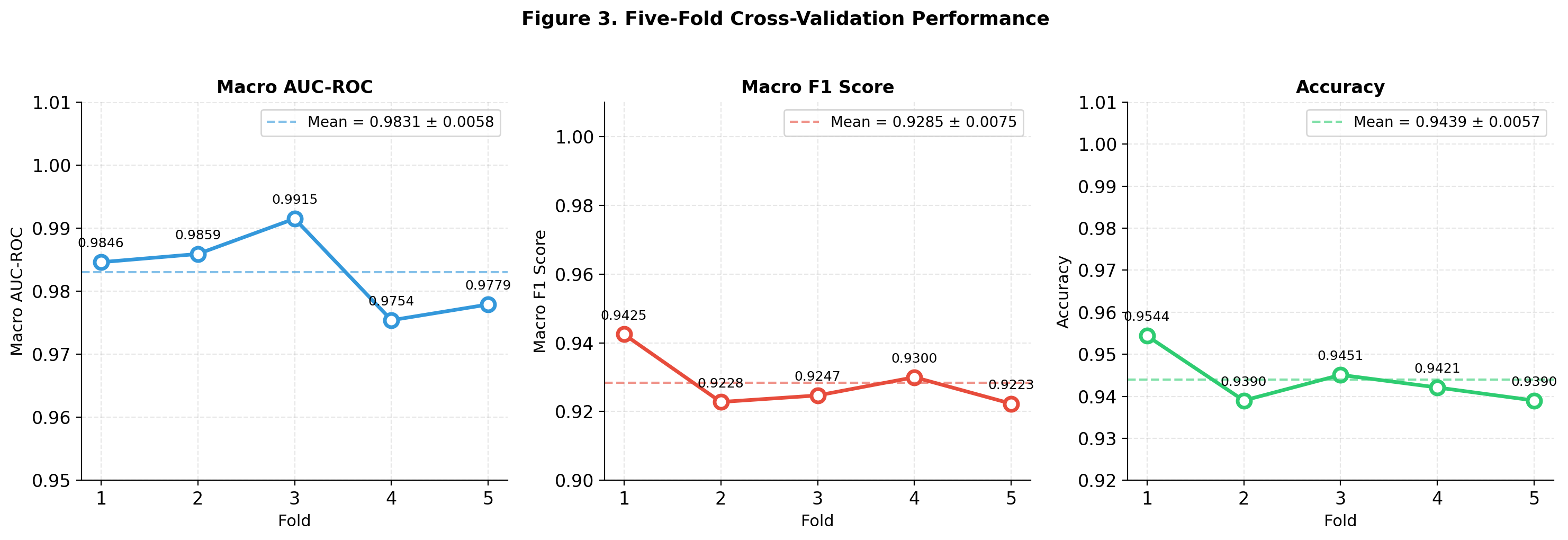}
\caption{Five-fold cross-validation results. Each point represents one fold. Dashed
horizontal line indicates the mean. AUC~=~Area Under the ROC Curve.}
\label{fig:fig3}
\end{figure}

On the held-out test set ($n=247$), the model achieved a macro AUC of 0.982
(95\% CI: 0.965--0.995), accuracy 0.943, balanced accuracy 0.932, macro F1
0.927, and Cohen's $\kappa$ 0.909. A $\kappa$ of 0.909 corresponds to
near-perfect agreement beyond chance~\cite{cohen1960}. Figure~\ref{fig:fig2}
presents ROC curves and confusion matrices. Figure~\ref{fig:fig3} shows
fold-wise cross-validation performance.

\begin{table}[H]
\centering
\caption{Comparison with published studies. AUC for this study is macro average
(one-versus-rest) with bootstrap 95\% CI based on 1,000~iterations. Results for
comparator studies taken from original publications.}
\label{tab:table2}
\rowcolors{2}{lightrow}{white}
\small
\begin{tabular}{p{3.2cm}p{2.2cm}ccccp{2.0cm}}
\toprule
\rowcolor{headerrow}
\textcolor{white}{\textbf{Study}} &
\textcolor{white}{\textbf{Method}} &
\textcolor{white}{\textbf{Classes}} &
\textcolor{white}{\textbf{AUC}} &
\textcolor{white}{\textbf{Accuracy}} &
\textcolor{white}{\textbf{XAI}} &
\textcolor{white}{\textbf{Data}} \\
\midrule
\rowcolor{highlightrow}
\textbf{This study (2026)} & XGBoost + Optuna & 3 & \textbf{0.982} & \textbf{94.3\%} & SHAP & 8 clinical \\
Yi et al.~\cite{yi2023} (2023)      & XGBoost-SHAP   & 3 & 0.91  & 87.6\%          & SHAP        & Clinical + MRI + APOE \\
Rashmi et al.~\cite{rashmi2025} (2025) & Grad. Boost & 2 & N/R   & 93.9\%          & SHAP        & Clinical (Kaggle) \\
Vlontzou et al.~\cite{vlontzou2025} (2025) & RF + SVM & 3 & N/R  & 87.5\% (BA)     & SHAP+LIME   & MRI + genetics \\
Akan et al.~\cite{akan2025} (2025)  & XGBoost        & 2 & 0.84  & N/R             & SHAP        & Clinical + biomarkers \\
Alatrany et al.~\cite{alatrany2024} (2024) & SVM     & 3 & N/R   & 90.7\% (F1)     & SHAP        & 1,024 features \\
Ritter et al.~\cite{ritter2015} (2015) & SVM         & 3 & 0.91  & 72.0\%          & None        & MRI + clinical \\
Kavitha et al.~\cite{kavitha2022} (2022) & RF        & 2 & 0.91  & 91.2\%          & None        & Clinical \\
Ding et al.~\cite{ding2019} (2019)  & Deep learning  & 2 & 0.96  & 82.4\%          & Grad-CAM    & PET scan \\
Battineni et al.~\cite{battineni2019} (2019) & SVM  & 2 & 0.85  & 84.7\%          & None        & Clinical \\
Zhang \& Shen~\cite{zhang2012} (2012) & Multi-task SVM & 3 & 0.87 & 79.3\%         & None        & MRI + PET \\
\bottomrule
\end{tabular}
\smallskip
\begin{minipage}{\linewidth}
\small\textit{BA = Balanced Accuracy; N/R = Not Reported; XAI = Explainability method;
MRI = Magnetic Resonance Imaging; PET = Positron Emission Tomography;
APOE = Apolipoprotein E gene; Grad-CAM = Gradient-weighted Class Activation Mapping.}
\end{minipage}
\end{table}

\subsection{SHAP Feature Importance}

\begin{figure}[H]
\centering
\includegraphics[width=\linewidth]{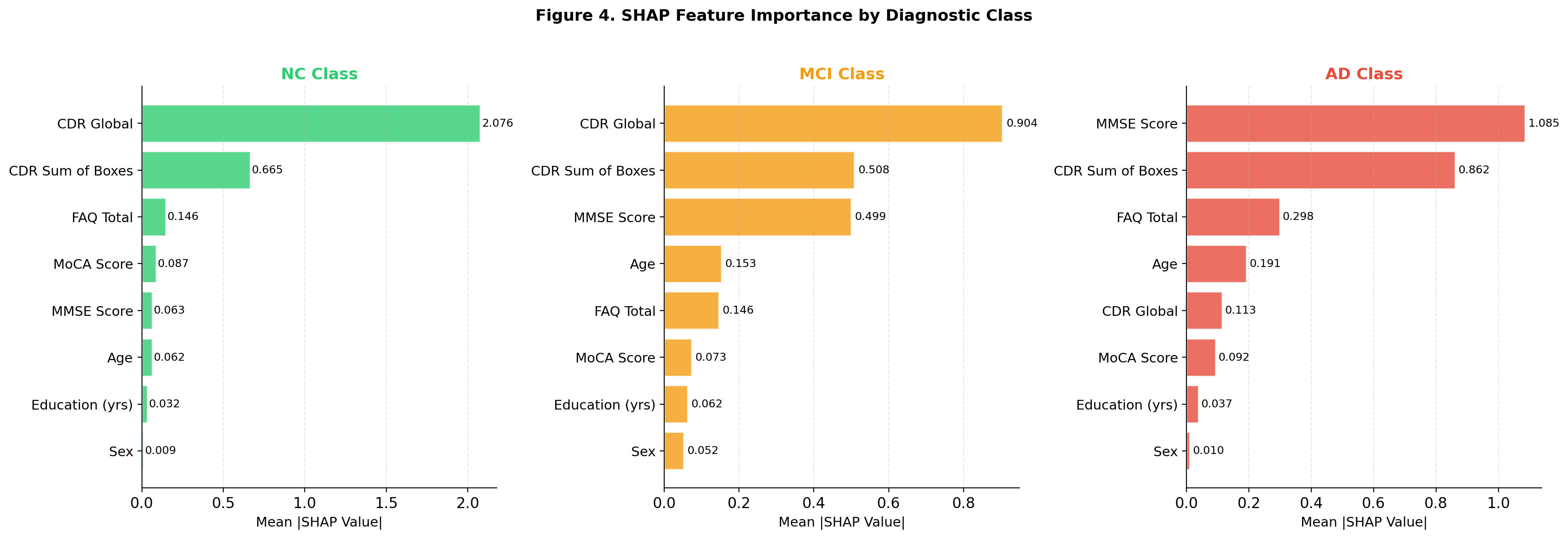}
\caption{SHAP feature importance by diagnostic class. Bar length represents mean absolute
SHAP value across all test set subjects. Higher values indicate greater importance.
Features are ranked in descending order.}
\label{fig:fig4}
\end{figure}

Table~\ref{tab:table3} presents the top five SHAP features for each diagnostic
class; Figure~\ref{fig:fig4} illustrates the full importance rankings.
Feature importance patterns were meaningfully differentiated across diagnostic
classes. For NC classification, CDR Global was overwhelmingly dominant
(mean~$|\text{SHAP}|=2.218$), followed distantly by CDR-SB (0.639).
For MCI classification, CDR Global remained the leading feature (1.417),
with MMSE score (0.463) and CDR-SB (0.322) jointly contributing.
For AD classification, CDR-SB (1.117) and MMSE score (0.942) were
co-dominant, with FAQ Total appearing consistently among the top predictors
for all three classes.

\begin{table}[H]
\centering
\caption{Top five SHAP features by diagnostic class (mean absolute SHAP value in parentheses).}
\label{tab:table3}
\rowcolors{2}{lightrow}{white}
\begin{tabular}{clll}
\toprule
\rowcolor{headerrow}
\textcolor{white}{\textbf{Rank}} &
\textcolor{white}{\textbf{NC class}} &
\textcolor{white}{\textbf{MCI class}} &
\textcolor{white}{\textbf{AD class}} \\
\midrule
1st & CDR Global (2.218) & CDR Global (1.417) & CDR-SB (1.117) \\
2nd & CDR-SB (0.639)     & MMSE Score (0.463) & MMSE Score (0.942) \\
3rd & FAQ Total (0.134)  & CDR-SB (0.322)     & FAQ Total (0.326) \\
4th & Age (0.107)        & FAQ Total (0.175)  & Age (0.263) \\
5th & MoCA (0.100)       & Age (0.140)        & CDR Global (0.077) \\
\bottomrule
\end{tabular}
\smallskip
\begin{minipage}{\linewidth}
\small\textit{SHAP = SHapley Additive exPlanations; CDR = Clinical Dementia Rating;
CDR-SB = CDR Sum of Boxes; MMSE = Mini-Mental State Examination;
FAQ = Functional Activities Questionnaire; MoCA = Montreal Cognitive Assessment.}
\end{minipage}
\end{table}

\section{Discussion}

This study demonstrates that an explainable XGBoost classifier trained on
eight routine clinical assessments can achieve near-perfect three-class
discrimination of NC, MCI, and AD, with a macro AUC of 0.982 on external
validation and a Cohen's $\kappa$ of 0.909. The closest methodologically
comparable study, \cite{yi2023}, also applied XGBoost-SHAP to the same
three-class ADNI task but required neuroimaging biomarkers and APOE genetic
data in addition to clinical assessments, achieving a lower AUC of 0.91 and
accuracy of 87.6\%. The present study surpasses that performance using only
eight routine clinical features, suggesting that neuroimaging and genetic data
may provide limited incremental discriminative value when clinical assessment
scores are comprehensively captured and optimally modelled.

\cite{rashmi2025} achieved comparable accuracy (93.9\%) using gradient
boosting on clinical and behavioural features, though that study used a
Kaggle-sourced dataset of unknown provenance rather than the standardised ADNI
cohort. \cite{vlontzou2025} achieved a balanced accuracy of 87.5\% on the
same three-class ADNI task using MRI volumetric measurements and genetic data;
the present study exceeds this (balanced accuracy 93.2\%) without any
neuroimaging requirement, further supporting the clinical utility of
assessment-only approaches. Deep learning models applied to MRI and PET scans
have achieved comparable or higher binary classification accuracy~\cite{ding2019},
but require expensive neuroimaging infrastructure unavailable in most clinical
settings and provide limited interpretability compared to SHAP-based
explanations.

The dominance of CDR Global in SHAP feature importance for both NC and MCI
classification reflects its well-established role as the primary clinical
staging instrument for dementia. The transition to CDR-SB dominance in AD
classification aligns with established evidence that CDR-SB provides greater
sensitivity for monitoring decline in moderate-to-severe AD~\cite{obryant2008},
as it captures the cumulative burden across all six CDR domains. The consistent
appearance of FAQ Total among top predictors for all three classes underscores
the clinical importance of functional assessment alongside cognitive testing,
a finding aligned with current diagnostic frameworks that require functional
impairment for an AD diagnosis~\cite{mckann2011}.

The differential SHAP importance patterns across diagnostic classes have
directly actionable clinical implications. For screening in primary care
settings where time is limited, CDR Global alone may be sufficient to identify
patients requiring specialist referral. For specialist memory clinics, the
combination of CDR-SB and MMSE provides the highest discriminative value for
confirming AD, while the subtler MCI profile, characterised by borderline CDR
Global with mild MMSE reduction, may benefit from additional cognitive testing
including MoCA. This staged assessment strategy could optimise clinical
resource allocation, particularly in lower-resource settings where
comprehensive neuropsychological batteries are not readily available.

MoCA had a 59\% missingness rate in this cohort, attributable to its
introduction in the ADNIGO phase and later. Despite this, MoCA appeared among
the top five predictors for NC classification, suggesting it provides
incremental information when available. The robustness of model performance
despite substantial MoCA missingness, managed through median imputation,
suggests that the remaining seven features capture sufficient information for
accurate classification.

Several limitations of the present study warrant acknowledgement. First, the
ADNI cohort is predominantly non-Hispanic White and highly educated (mean
15.0--16.5~years), which may limit generalisability to ethnically diverse
populations, including Middle Eastern and South Asian communities where genetic
risk profiles and educational norms differ substantially~\cite{elhaayek2019}.
Validation in diverse external cohorts is a priority. Second, the analysis is
cross-sectional using baseline data only; incorporating longitudinal cognitive
trajectories may further improve MCI conversion prediction. Third, no
comparison was made against commercial AI tools or structured clinical
algorithms, which would be informative for contextualising the model's
diagnostic utility. Fourth, while SHAP values provide rigorous post-hoc
explanations, they reflect statistical associations rather than causal
mechanisms.

Future work will extend this pipeline in two directions. First,
speech-language biomarkers will be incorporated, extracted from spontaneous
speech recordings including acoustic features (MFCCs, prosody, pause patterns)
and linguistic features (type-token ratio, lexical richness, syntactic
complexity), using the DementiaBank Pitt Corpus and ADReSS challenge datasets.
A multimodal attention fusion model combining clinical assessment features with
speech features could be deployed via a telephone call, enabling screening in
settings where neuropsychological testing is inaccessible. Second, validation
on a Saudi Arabian clinical cohort is planned, where both incidence of dementia
and AI research infrastructure are rapidly growing, addressing a critical equity
gap in global dementia AI literature.

\section{Conclusion}

An explainable XGBoost classifier was developed and externally validated for
three-class early detection of Alzheimer's disease using routine clinical
assessments from the ADNI dataset. The model achieved a macro AUC of 0.982
(95\% CI: 0.965--0.995) and Cohen's $\kappa$ of 0.909, with SHAP analysis
confirming clinically meaningful, class-specific feature importance patterns.
CDR Global dominated NC and MCI classification, while CDR-SB and MMSE together
drove AD detection, findings that align with established diagnostic criteria
and support the model's clinical validity. These results demonstrate that
highly accurate and interpretable Alzheimer's detection is achievable from
assessments routinely performed in memory clinics worldwide, supporting the
feasibility of scalable AI-assisted screening, particularly in
resource-limited settings.

\section*{Acknowledgements}

Data collection and sharing for this project was funded by the Alzheimer's
Disease Neuroimaging Initiative (ADNI; National Institutes of Health Grant
U01~AG024904) and DOD ADNI (Department of Defense award number
W81XWH-12-2-0012). ADNI is funded by the National Institute on Aging, the
National Institute of Biomedical Imaging and Bioengineering, and through
generous contributions from the following: AbbVie; Alzheimer's Association;
Alzheimer's Drug Discovery Foundation; Araclon Biotech; BioClinica, Inc.;
Biogen; Bristol-Myers Squibb Company; CereSpir, Inc.; Cognitect; Eisai Inc.;
Elan Pharmaceuticals, Inc.; Eli Lilly and Company; EuroImmun; F.~Hoffmann-La
Roche Ltd and its affiliated company Genentech, Inc.; Fujirebio; GE Healthcare;
IXICO Ltd.; Janssen Alzheimer Immunotherapy Research \& Development, LLC;
Johnson \& Johnson Pharmaceutical Research \& Development LLC; Lumosity;
Lundbeck; Merck \& Co., Inc.; Meso Scale Diagnostics, LLC; NeuroRx Research;
Neurotrack Technologies; Novartis Pharmaceuticals Corporation; Pfizer Inc.\ and
Janssen Alzheimer Immunotherapy Research \& Development, LLC; Servier; Takeda
Pharmaceutical Company; and Transition Therapeutics.

\section*{Conflicts of Interest}

The author declares no conflicts of interest. The funders had no role in the
design of the study, data collection and analysis, decision to publish, or
preparation of the manuscript.

\section*{Data Availability}

The ADNI dataset is publicly available to registered users at
\url{https://adni.loni.usc.edu}. The analysis code and trained model are
available at \url{https://github.com/[to-be-added-upon-acceptance]}.

\bibliographystyle{unsrt}
\bibliography{references}

\end{document}